\documentclass[conference]{IEEEtran}
\usepackage{graphicx}
\usepackage{amsmath}
\usepackage{cite}
\usepackage[hidelinks]{hyperref} 

\title{Enhancing Multi-Label Thoracic Disease Diagnosis with Deep Ensemble-Based Uncertainty Quantification}

\author{
    \IEEEauthorblockN{Yasiru Laksara}
    \IEEEauthorblockA{Department of Computer Science and Engineering \\
    University of Moratuwa \\
    Katubedda 10400, Sri Lanka \\
    yasirul.21@cse.mrt.ac.lk}
    \and
    \IEEEauthorblockN{Uthayasanker Thayasivam}
    \IEEEauthorblockA{Department of Computer Science and Engineering \\
    University of Moratuwa \\
    Katubedda 10400, Sri Lanka \\
    rtuthaya@cse.mrt.ac.lk}
}

\begin{document}

\maketitle

\begin{abstract}
The utility of deep learning models, such as CheXNet, in high stakes clinical settings is fundamentally constrained by their purely deterministic nature, failing to provide reliable measures of predictive confidence. This project addresses this critical gap by integrating robust Uncertainty Quantification (UQ) into a high performance diagnostic platform for 14 common thoracic diseases on the NIH ChestX-ray14 dataset. Initial architectural development failed to stabilize performance and calibration using Monte Carlo Dropout (MCD), yielding an unacceptable Expected Calibration Error (ECE) of 0.7588. This technical failure necessitated a rigorous architectural pivot to a high diversity, 9-member Deep Ensemble (DE). This resulting DE successfully stabilized performance and delivered superior reliability, achieving a State-of-the-Art (SOTA) average Area Under the Receiver Operating Characteristic Curve (AUROC) of 0.8559 and an average F1 Score of 0.3857. Crucially, the DE demonstrated superior calibration (Mean ECE of 0.0728 and Negative Log-Likelihood (NLL) of 0.1916) and enabled the reliable decomposition of total uncertainty into its Aleatoric (irreducible data noise) and Epistemic (reducible model knowledge) components, with a mean Epistemic Uncertainty (EU) of 0.0240. These results establish the Deep Ensemble as a trustworthy and explainable platform, transforming the model from a probabilistic tool into a reliable clinical decision support system.
\end{abstract}


\section{Introduction}

Deep learning has achieved remarkable diagnostic accuracy in medical imaging, exemplified by the CheXNet model, which reached radiologist-level performance for pneumonia detection and classification across 14 thoracic pathologies. Despite these advances, clinical adoption of such models remains limited. A primary barrier is their reliance on deterministic, single point predictions that lack a quantifiable measure of confidence. In high-stakes environments such as medical diagnosis, an overconfident yet incorrect prediction can have severe consequences. Clinicians therefore require systems that not only provide accurate results but also indicate when a case is ambiguous, noisy, or outside the model’s knowledge domain. Uncertainty Quantification (UQ) provides this crucial layer of trust, enabling models to flag high-risk or uncertain cases for expert review and thereby bridging the gap between algorithmic accuracy and clinical reliability.

\subsection{Architectural Challenges and Strategic Pivot}

The project was initially conceived to enhance the CheXNet base model (DenseNet-121) by incorporating Monte Carlo Dropout (MCD), chosen for its computational efficiency as an approximation to Bayesian Neural Networks. However, this approach proved technically infeasible. Early R\&D confirmed the well-documented challenges in reproducing the original CheXNet paper’s metrics on public test sets, prompting the selection of the more stable and reproducible {DannyNet} implementation also based on DenseNet-121 as the experimental baseline \href{https://doi.org/10.48550/arXiv.2505.06646}{[1]}.

Attempts to integrate UQ through MCD on this stabilized baseline resulted in significant degradation of classification performance and severe predictive miscalibration. The MCD model yielded an alarmingly high Expected Calibration Error (ECE) of 0.7588, indicating a poor correspondence between predicted confidence and actual accuracy. This quantitative failure highlighted that MCD’s limited stochastic sampling explores only a restricted region of the loss landscape, making it inadequate for the complexity of multi-label thoracic disease classification. Consequently, a strategic architectural pivot was undertaken toward the Deep Ensemble (DE) method, which is computationally intensive but well established for delivering robust and state-of-the-art uncertainty estimation.

\subsection{Contributions}

The successful implementation and deployment of the Deep Ensemble architecture yielded three primary contributions:

\begin{itemize}
    \item \textbf{High-Performance UQ Platform:} Development of a 9-member Deep Ensemble composed of diverse architectural backbones (DenseNet, EfficientNet, and CBAM-enhanced variants) and advanced multi-label loss functions (Focal Loss and ZLPR Loss). The final ensemble achieved a \textbf{state-of-the-art average AUROC of 0.8559} in our experimental setup.
    \item \textbf{Validated Uncertainty and Calibration:} Rigorous evaluation demonstrated excellent calibration ({ECE = 0.0728}) and enabled a reliable decomposition of total uncertainty into {Aleatoric} (irreducible data noise) and {Epistemic} (reducible model knowledge) components, with a mean Epistemic Uncertainty (EU) of {0.0240}.
    \item \textbf{Interpretable Decision Making:} Integration of {Ensemble Grad-CAM} visualizations, which generate consensus driven heatmaps by averaging feature importance across ensemble members, thereby enhancing transparency and interpretability in clinical decision support.
\end{itemize}

Together, these contributions advance the reliability, interpretability, and clinical readiness of deep learning based thoracic disease classifiers.

\section{Related Work}
\subsection{Benchmarking in Thoracic Disease Classification}

The foundational benchmark in thoracic disease classification is CheXNet, a DenseNet-121--based model trained to classify 14 thoracic diseases \cite{rajpurkar2017chexnet}. While CheXNet achieved radiologist-level performance, reproducing its published metrics on public test sets has proven challenging, highlighting a critical reproducibility issue in medical machine learning research \cite{strick2025reproducing}.

To address this limitation, this study adopts DannyNet, which shares the DenseNet-121 backbone but has been empirically validated as a more robust and reproducible baseline on the NIH ChestX-ray14 dataset. This dataset contains 112,120 frontal chest X-rays from 30,805 patients, labeled via NLP extraction from radiology reports with $\approx$90\% accuracy. Its class imbalance (e.g., rare Hernia versus common Infiltration) and multi-label co-occurrence patterns make uncertainty aware modeling essential for reliable clinical predictions \cite{strick2025reproducing, nihchestxray2018kaggle}.

\subsection{Theoretical Framework and Importance of Uncertainty}
As established in the introduction, quantifying uncertainty is crucial for the clinical adoption of AI models in medical imaging, moving beyond raw accuracy to measure predictive confidence. A model's total uncertainty can be broken down into two main types:

\textbf{Aleatoric Uncertainty (AU):} This is the inherent, irreducible noise in the data itself. In medical imaging, this can be caused by factors like sensor noise, imaging artifacts, or, critically, ambiguous labels from the source radiology reports, which often use phrases like ``could be due to'' or ``cannot be excluded.'' A model cannot reduce this type of uncertainty by seeing more data \cite{kendall2017bayesian}.

\textbf{Epistemic Uncertainty (EU):} This is the model's own uncertainty due to a lack of knowledge or limited data. It can be reduced by providing the model with more data and is often high for out-of-distribution (OOD) or ``never-before-seen'' samples \cite{kendall2017bayesian}.

A key finding from recent research is that incorporating uncertainty labels during model training leads to higher predictive variance for uncertain cases at test time, preventing the model from making ``over-confident mistakes'' \cite{kendall2017bayesian}. This ability is of significant clinical value, particularly when a model flags an out-of-distribution case that a clinician should review. The field of UQ for medical imaging is a growing area of research, with recent reviews providing a comprehensive overview of both probabilistic and non-probabilistic methods.

\subsection{General Approaches in Uncertainty Quantification}
The field of Uncertainty Quantification in deep learning has seen a variety of approaches, spanning both probabilistic and non-probabilistic methods, particularly in medical image analysis. These studies often categorize methods into ``distributional'' and ``deterministic'' approaches. Distributional methods, such as Latent Heteroscedastic Classifiers, model uncertainty by learning a second order predictive distribution, where the output is a distribution over predictions rather than a single point estimate. This category includes a wide array of techniques, such as Stochastic Weight Averaging in Parallel (SWAG), Evidential Deep Learning (EDL), and the Deep and Shallow Ensembles we explore \cite{baur2025benchmarking}. In contrast, deterministic methods, like Loss Prediction or Temperature Scaling, estimate uncertainty without requiring a predictive distribution, often relying on model-internal features to assess confidence. Other research has also explored the information-theoretical approach for disentangling epistemic and aleatoric uncertainty, though recent studies have raised concerns about its practical effectiveness in complex, real-world datasets like those used in medical imaging \cite{chan2025estimating}.

\subsubsection{\textbf{Probabilistic Methods: Bayesian Deep Learning and Approximations}}

Bayesian deep learning (BDL) is a promising approach for uncertainty quantification in healthcare as it provides a probabilistic framework that quantifies uncertainty and enhances prediction reliability.

\textit{Full Bayesian Neural Networks (BNN) and MCMC:} Directly implementing a full Bayesian Neural Network (BNN) using methods like Markov Chain Monte Carlo (MCMC) is often computationally intractable for large, modern neural networks due to the high dimensional parameter space. MCMC methods, while theoretically sound, produce autocorrelated samples and can be inefficient in exploring the target distribution, leading to challenges with convergence and high computational costs \cite{papamarkou2022challenges}.

\textit{Variational Inference (VI):} This method approximates the true posterior distribution of the model's weights with a simpler, more tractable distribution, such as a Gaussian distribution. The goal is to make the approximate distribution as close as possible to the true posterior by minimizing a measure like the Kullback-Leibler (KL) divergence. While VI offers a principled Bayesian framework, it still has limitations and may not fully capture the complex uncertainty landscape \cite{margossian2025variational}.

\textit{Monte Carlo Dropout (MCD):} This is a practical and computationally efficient approximation of a BNN that leverages dropout, a technique traditionally used for regularization, by keeping it active during inference. For a single input image, multiple forward passes are performed with different neurons randomly "dropped out," effectively creating an ensemble of "random subnets" \cite{hasan2022controlled}. The final prediction is the average of these passes, and the uncertainty is quantified by the variance of the predictions. While more efficient than full Bayesian methods, MCD tends to sample from a single, slightly varied region of the loss landscape, yielding a "distribution of similar functions." These limitations have been further explored in recent works \cite{aws2025deepensembles, whata2024uncertainty}.

\textit{Deep Ensembles (DE):} Deep Ensembles offer a more robust, albeit more computationally intensive, approach to UQ that is often considered a non-Bayesian approach, although some have explored its links to Bayesian methods. The method involves training a collection of identical models with different random initializations. The diversity in initial weight values and training trajectories causes the models to converge to different, well separated "low-loss valleys" in the parameter space, leading to a "distribution of diverse functions" \cite{aws2025deepensembles}. This diversity makes Deep Ensembles a state-of-the-art method for uncertainty quantification, particularly in out-of-distribution settings, where they have been shown to outperform MCD and other methods. However, the computational cost of training multiple full models is a significant limitation, which researchers are trying to address with more efficient ensemble methods like Stochastic Weight Averaging in Parallel (SWAP) \cite{lee2025advanceduncertainty}.

\textit{Ensemble Bayesian Neural Networks (EBNN):} This is a specific type of ensemble rooted in Bayesian theory. EBNN addresses epistemic uncertainty by learning a posterior distribution over the model's parameters. It involves sampling and weighting networks according to this posterior to form an ensemble model, also referred to as a "Bayes ensemble." This is conceptually distinct from a standard Deep Ensemble, as it leverages the probabilistic framework of BNNs. While theoretically sound, EBNN is often computationally demanding due to the complexity of sampling from the posterior distribution, and its performance may not always surpass simpler uniformly weighted deep ensembles, according to recent research \cite{whata2024uncertainty, lakshminarayanan2017simple}.

\textit{Ensemble Monte Carlo (EMC) Dropout:} This is an extension of Monte Carlo Dropout designed to overcome its limitations. One recent approach proposes a strategy to compute an ensemble of subnetworks, each corresponding to a non-overlapping dropout mask, and trains them independently via a pruning strategy. The goal of this method is to bridge the performance gap between MC Dropout and Deep Ensembles, achieving similar accuracy and uncertainty estimates to deep ensembles while maintaining the computational efficiency of MC Dropout \cite{whata2024uncertainty}.

\subsubsection{\textbf{Non-Probabilistic Methods}}

In contrast to probabilistic methods that model a predictive distribution, non-probabilistic or deterministic methods estimate uncertainty based on internal model properties without a rigorous statistical framework.

\textit{Conformal Prediction (CP):} This is a prominent non-probabilistic method that provides statistically rigorous prediction sets for each prediction. CP guarantees that the true label will be included in the prediction set at a user specified error rate on average across the entire test distribution. However, this guarantee is marginal and does not hold for specific subgroups or individual data points, which can be a significant limitation in clinical applications where rare but critical classes may be systematically under covered \cite{epistemic2025singlemodel}.

\textit{Temperature Scaling:} This is a post-hoc calibration technique that adjusts a model's confidence scores by dividing the logits (model outputs before the softmax layer) by a single scalar value called the "temperature." This method has been shown to improve the calibration of a model's confidence but is limited as the temperature is calculated on a validation set and may not generalize well to out-of-distribution data \cite{guo2017calibration, zeng2025uncertainty}.

\subsection{Multi-Label Classification Loss Functions}
\label{sec:multi_label_loss_functions}
To effectively tackle the challenges presented by the NIH ChestX-ray14 dataset, particularly the class imbalance and label co-occurrence, the systematic search incorporated two specialized loss functions:

\textbf{Focal Loss:} This standard technique mitigates class imbal-
ance by assigning lower weights to easily classified negative
examples, forcing the model to focus training efforts on rare
or hard to classify samples \cite{strick2025reproducing, kunang2021deeplearning}.

\textbf{ZLPR Loss (Zero-threshold Log-sum-exp Pairwise Ranking Loss):} This loss function was explicitly included to address complex multi-label classification issues, such as uncertain target label counts and inter label correlations, which are prevalent in the NIH ChestX-ray14 dataset \cite{su2022zlpr}.

\section{Methodology}

\begin{figure}[!t]
    \centering
    \includegraphics[width=\linewidth]{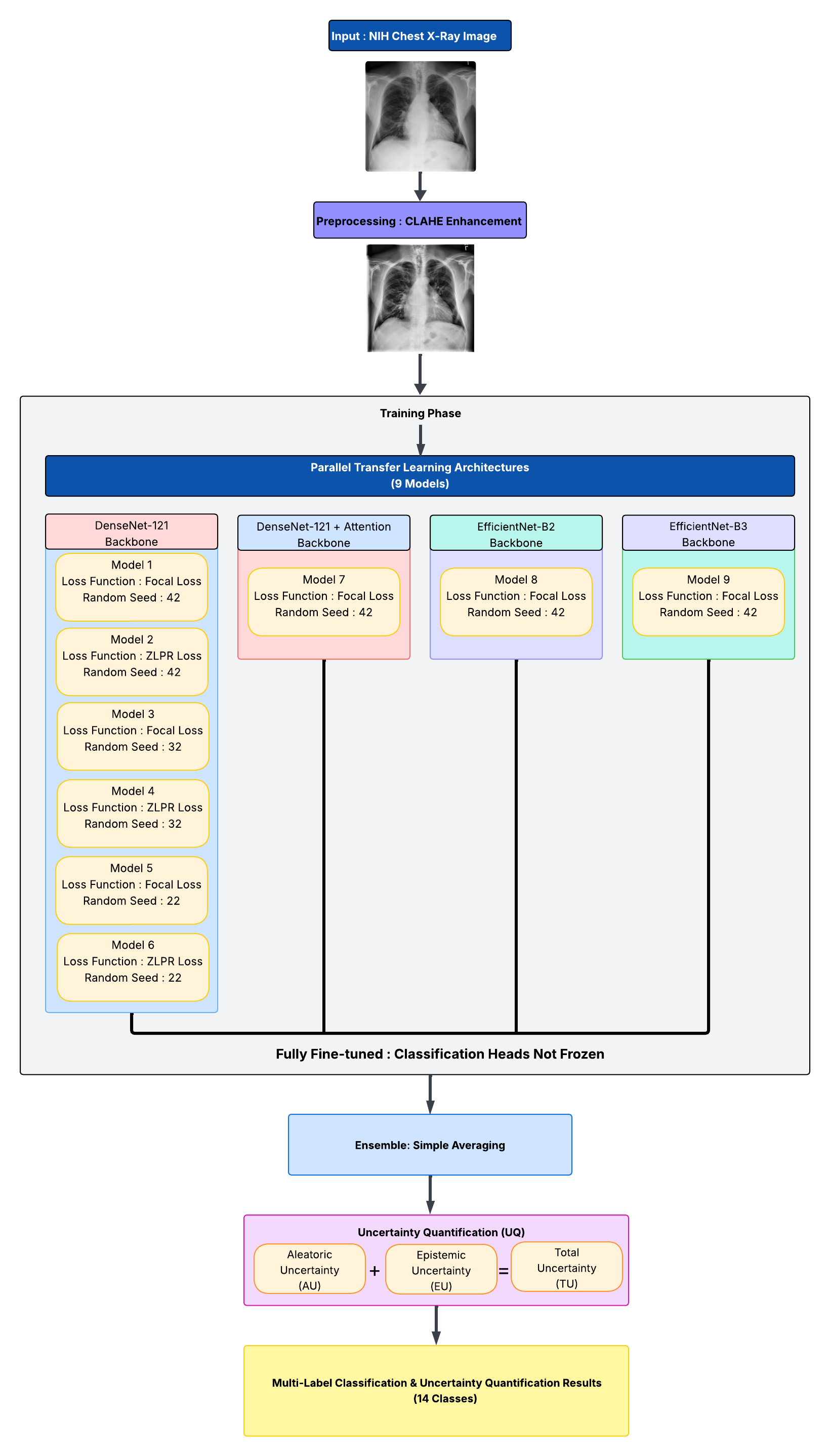}
    \caption{Overview of the methodology pipeline}
    \label{fig:methodology_pipeline}
\end{figure}

The final architectural solution, detailed in Figure~\ref{fig:methodology_pipeline}, was the result of a systematic, multi-stage Research and Development process designed to maximize predictive power and ensure a stable platform for Uncertainty Quantification (UQ).

\subsection{Data Preparation and Patient-Level Splitting}
\label{sec:data_preparation}
All experiments were conducted using the NIH Chest X-ray14 multi-label dataset. To ensure an unbiased evaluation of generalization performance and prevent patient data leakage, the dataset was strictly split at the patient ID level. This procedure reserved 2\% of unique patients for the final, held-out test set, and allocated 5.2\% of the remaining patients to the validation set. This stratification guaranteed that no images from the same individual appeared in more than one data split.

\textbf{Preprocessing:} Beyond standard normalization to ImageNet statistics, a crucial addition to the baseline preprocessing protocol was the application of {Contrast Limited Adaptive Histogram Equalization (CLAHE)}. CLAHE was utilized to enhance local contrast and improve the visibility of subtle chest structures prior to model input.

\textbf{Input Sizes:} Input images were resized according to the specific requirements of the chosen backbone architectures for compatibility with pre-trained ImageNet weights: 224×224 pixels for DenseNet-121 and its attention enhanced variant, 260×260 pixels for EfficientNet-B2, and 300×300 pixels for EfficientNet-B3.

\subsection{Base Model Training Configuration}
All base models were initialized with ImageNet-pretrained weights and trained using a \textbf{full fine-tuning} strategy on the NIH ChestX-ray14 dataset, where no layers were frozen and all parameters were updated end-to-end. This approach allowed the model backbones to fully adapt to the domain specific characteristics of chest X-rays.

\textbf{Training Parameters:} Models were trained for a maximum of 25 epochs with early stopping (patience =5) applied if validation loss did not improve. Training employed the AdamW optimizer ($\beta_1=0.9$, $\beta_2=0.999$, $\epsilon=1\times10^{-8}$) with a base learning rate of $5\times10^{-5}$ and weight decay of $1\times10^{-5}$. A ReduceLROnPlateau scheduler monitored validation loss, reducing the learning rate by a factor of 0.1 after one epoch without improvement.

\textbf{Loss Functions:} To manage the severe class imbalance and multi-label complexity of the dataset, two advanced loss functions were employed: Focal Loss and ZLPR Loss, as detailed in Section \ref{sec:multi_label_loss_functions}.

\subsection{Systematic Model Search and Architectural Pivot Rationale}
The decision to pivot to a Deep Ensemble was motivated by the quantitative failure of Monte Carlo Dropout (MCD) to maintain performance integrity, as detailed in Section \ref{sec:overall_classification_performance}.

\textbf{Systematic Search:} A pool of 14 distinct models was generated through a systematic search by varying conditions across three primary axes to maximize diversity:

\textbf{Architectural Diversity:} Exploration included the DenseNet-121 backbone (the CheXNet/DannyNet base), an attention-enhanced variant (DenseNet-121 with CBAM), EfficientNet-B2, and EfficientNet-B3.

\textbf{Loss Function Diversity:} Models were trained using both Focal Loss ($\alpha = 1$, $\gamma = 2$) and ZLPR Loss.

\textbf{Initialization Diversity:} Training utilized different random seeds (22, 32, 42) to ensure convergence to genuinely diverse regions in the parameter space.

\textbf{MCD Implementation:} The initial uncertainty quantification (UQ) trial applied Monte Carlo Dropout (MCD) to the DenseNet-121 model. A dropout layer was inserted immediately before the final classification layer, and inference was performed using $T = 30$ stochastic forward passes, where $T$ denotes the number of Monte Carlo samples used to estimate predictive variance. The empirical results of this trial, presented in Section \ref{sec:overall_classification_performance}, revealed that the MCD approximation was insufficiently robust for this multi-label classification task, motivating the adoption of the Deep Ensemble approach.

\subsection{Deep Ensemble Construction and Uncertainty Formulation}
\textbf{Model Selection:} Based on rigorous evaluation of individual test AUROC, F1 score, and architectural complementarity across the 14 trials, the 9 highest performing and most diverse models were selected to form the final Deep Ensemble ($M = 9$). The components of this ensemble are listed in Table \ref{tab:ensemble_components_final} (Section \ref{sec:experiments_and_results}).

\textbf{Classification Output:} Let $M$ be the number of ensemble members, $k$ index the classes, and $p_k^{(m)}$ denote the predicted probability of class $k$ by the $m$-th model. The ensemble’s final prediction $\hat{p}(y|x)$ is obtained by a uniform average over all members:

\begin{equation}
\bar{p}_k = \frac{1}{M} \sum_{m=1}^{M} p_k^{(m)}
\end{equation}

\textbf{Uncertainty Decomposition:} The final UQ is formulated by decomposing the Total Uncertainty (TU) into its constituent components, Aleatoric Uncertainty (AU) and Epistemic Uncertainty (EU), using the standard framework:

\textbf{Total Uncertainty (TU):}
\begin{equation}
TU = -\sum_{k} \bar{p}_k \log \bar{p}_k
\end{equation}

\textbf{Aleatoric Uncertainty (AU):}
\begin{equation}
AU = \frac{1}{M} \sum_{m=1}^{M} \left(-\sum_{k} p_k^{(m)} \log p_k^{(m)} \right)
\end{equation}

\textbf{Epistemic Uncertainty (EU):}
\begin{equation}
EU = TU - AU
\end{equation}

Here, TU measures the overall uncertainty in the ensemble’s prediction, AU captures the irreducible uncertainty inherent in the data, and EU quantifies the uncertainty arising from the model’s lack of knowledge. This decomposition enables more interpretable predictions and informs when expert review may be necessary.

\section{Experiments and Results}
\label{sec:experiments_and_results}

\begin{table*}[t]
    \centering
    \caption{Deep Ensemble Model Components and Individual Test Performance}
    \resizebox{\textwidth}{!}{
    \begin{tabular}{lllll}
        \hline
        Model Name & Backbone/Loss Function & Diversity Factor & Test AUROC & Test F1 \\
        \hline
        seed 42 DenseNet-121 - Focal Loss & DenseNet-121 / Focal & CLAHE Enhanced SOTA Baseline & \textbf{0.8514} & 0.3803 \\
        seed 22 DenseNet-121 - Focal Loss & DenseNet-121 / Focal & Random Seed (22) & 0.8475 & \textbf{0.3852} \\
        seed 32 DenseNet-121 - Focal Loss & DenseNet-121 / Focal & Random Seed (32) & 0.8458 & 0.3679 \\
        seed 42 DenseNet-121 + Attention Focal Loss & DenseNet-121+CBAM / Focal & Architecture (CBAM) & 0.8480 & 0.3787 \\
        seed 42 Efficient Net-B2 Focal Loss & EfficientNet-B2 / Focal & Architecture (EffNet-B2) & 0.8322 & 0.3528 \\
        seed 42 Efficient Net-B3 Focal Loss & EfficientNet-B3 / Focal & Architecture (EffNet-B3) & 0.8117 & 0.3338 \\
        seed 22 DenseNet-121 - ZLPR Loss & DenseNet-121 / ZLPR & Loss Function (ZLPR) + Seed (22) & 0.8468 & 0.3758 \\
        seed 32 DenseNet-121 - ZLPR Loss & DenseNet-121 / ZLPR & Loss (ZLPR) + Seed (32) & 0.8479 & 0.3762 \\
        seed 42 DenseNet-121 ZLPR Loss & DenseNet-121 / ZLPR & Loss Function (ZLPR) & 0.8462 & 0.3621 \\
        \hline
    \end{tabular}}
    \vspace{0.5em}
    \footnotesize \textit{Note: All models were trained using CLAHE-enhanced input images as described in Section \ref{sec:data_preparation}.}
    \label{tab:ensemble_components_final}
\end{table*}

\begin{table*}[t]
    \centering
    \caption{Overall Performance Comparison and R\&D Validation}
    \resizebox{\textwidth}{!}{
    \begin{tabular}{llllll}
        \hline
        Metric & CheXNet (Paper) & DannyNet (Paper) SOTA & DannyNet (Reproduced) & MCD Trial Result & \begin{tabular}[c]{@{}c@{}}Deep Ensemble (DE) \\ (Our Work)\end{tabular} \\
        \hline
        Avg AUROC & 0.8066 & 0.8527 & 0.8471 & 0.8362 & \textbf{0.8559} \\
        Avg F1 Score & 0.435 & 0.3861 & 0.3705 & 0.3713 & \textbf{0.3857} \\
        Loss & — & 0.0416 & 0.0419 & 0.0426 & \textbf{N/A} \\
        \hline
    \end{tabular}}
    \label{tab:overall_performance_final}
\end{table*}

\begin{table*}[t]
    \centering
    \caption{Detailed Per-Class UQ and Calibration Results for Deep Ensemble}
    \resizebox{\textwidth}{!}{
    \begin{tabular}{llllllllll}
        \hline
        Disease & AUROC & F1 Score & Threshold & Brier Score & ECE & NLL & TU Mean & AU Mean & EU Mean \\
        \hline
        Atelectasis & 0.8215 & 0.4084 & 0.3153 & 0.0841 & 0.0953 & 0.3037 & 0.4512 & 0.4267 & 0.0245 \\
        Cardiomegaly & 0.9405 & 0.5035 & 0.2786 & 0.0383 & 0.0508 & 0.1527 & 0.2716 & 0.2467 & 0.0249 \\
        Consolidation & 0.7834 & 0.2402 & 0.2115 & 0.0476 & 0.0868 & 0.2053 & 0.3664 & 0.3418 & 0.0246 \\
        Edema & 0.8976 & 0.2635 & 0.2556 & 0.0243 & 0.0570 & 0.1165 & 0.2448 & 0.2235 & 0.0213 \\
        Effusion & 0.9059 & 0.6245 & 0.3656 & 0.0885 & 0.0922 & 0.3062 & 0.4444 & 0.4209 & 0.0235 \\
        Emphysema & 0.9705 & 0.5584 & 0.2631 & 0.0208 & 0.0567 & 0.1057 & 0.2404 & 0.2165 & 0.0239 \\
        Fibrosis & 0.8448 & 0.1531 & 0.2368 & 0.0212 & 0.0679 & 0.1172 & 0.2687 & 0.2432 & 0.0255 \\
        Hernia & 0.9937 & 0.7500 & 0.2836 & 0.0021 & 0.0195 & 0.0241 & 0.0937 & 0.0805 & 0.0133 \\
        Infiltration & 0.7078 & 0.4145 & 0.3081 & 0.1402 & 0.1006 & 0.4524 & 0.5660 & 0.5467 & 0.0193 \\
        Mass & 0.9126 & 0.4917 & 0.3270 & 0.0430 & 0.0873 & 0.1877 & 0.3570 & 0.3287 & 0.0283 \\
        Nodule & 0.7862 & 0.3258 & 0.2939 & 0.0584 & 0.0865 & 0.2391 & 0.3974 & 0.3698 & 0.0276 \\
        Pleural Thickening & 0.8073 & 0.2448 & 0.2296 & 0.0425 & 0.0736 & 0.1861 & 0.3357 & 0.3089 & 0.0268 \\
        Pneumonia & 0.7193 & 0.0683 & 0.1853 & 0.0183 & 0.0710 & 0.1150 & 0.2784 & 0.2524 & 0.0260 \\
        Pneumothorax & 0.8908 & 0.3537 & 0.3422 & 0.0395 & 0.0747 & 0.1704 & 0.3217 & 0.2957 & 0.0260 \\
        \hline
    \end{tabular}}
    \label{tab:detailed_per_class_results}
\end{table*}

\subsection{Evaluation Metrics}
The evaluation protocol assessed performance across three domains: classification accuracy (AUROC, F1 Score), predictive reliability (ECE, NLL, Brier Score), and uncertainty quality (TU, AU, EU). AUROC served as the primary metric due to the multi-label nature and severe class imbalance of the dataset. ECE, NLL, and Brier Score quantified the alignment between predicted probabilities and true correctness.

\subsection{Overall Classification Performance and MCD Failure Validation}
\label{sec:overall_classification_performance}
The R\&D process required a strategic pivot validated by the catastrophic performance of the initial UQ approach. The first attempt to integrate Monte Carlo Dropout (MCD) resulted in an unacceptable decline in core classification performance and catastrophic predictive miscalibration relative to the deterministic DannyNet baseline.

Table \ref{tab:overall_performance_final} summarizes the architectural R\&D outcomes. The MCD trial yielded an average AUROC of 0.8362 and an average F1 score of 0.3713, showing a minor degradation compared to the reproduced DannyNet baseline (AUROC 0.8471, F1 0.3705). Crucially, the MCD model exhibited an alarming lack of calibration, with the detailed metrics shown in Table \ref{tab:calibration_metrics_final_new}. The Expected Calibration Error (ECE) of 0.7588 confirmed its technical non-viability for clinical application.

\begin{table}[!h]
    \centering
    \caption{Summary of Mean Uncertainty and Calibration Metrics for Deep Ensemble}
    \resizebox{0.95\columnwidth}{!}{
    \begin{tabular}{lllll}
        \hline
        Uncertainty Type & Mean Value & Metric & Mean Value \\
        \hline
        Total Uncertainty (TU) & 0.3312 & Brier Score & 0.0478 \\
        Aleatoric Uncertainty (AU) & 0.3073 & ECE & 0.0728 \\
        Epistemic Uncertainty (EU) & 0.0240 & NLL & 0.1916 \\
        \hline
    \end{tabular}}
    \label{tab:Summary_of_Mean_Uncertainty_and_Calibration_Metrics_for_Deep_Ensemble}
\end{table}

\begin{table}[h]
    \centering
    \caption{Calibration Metrics: Failure Analysis of Monte Carlo Dropout (MCD)}
    \begin{tabular}{ll}
        \hline
        Metric & MCD Trial Result \\
        \hline
        Negative Log-Likelihood (NLL) & 0.2526 \\
        Expected Calibration Error (ECE) & 0.7588 \\
        Brier Score & 0.0631 \\
        \hline
    \end{tabular}
    \label{tab:calibration_metrics_final_new}
\end{table}

\begin{figure}[t]
    \centering
    \includegraphics[width=\columnwidth,height=0.5\textheight,keepaspectratio]{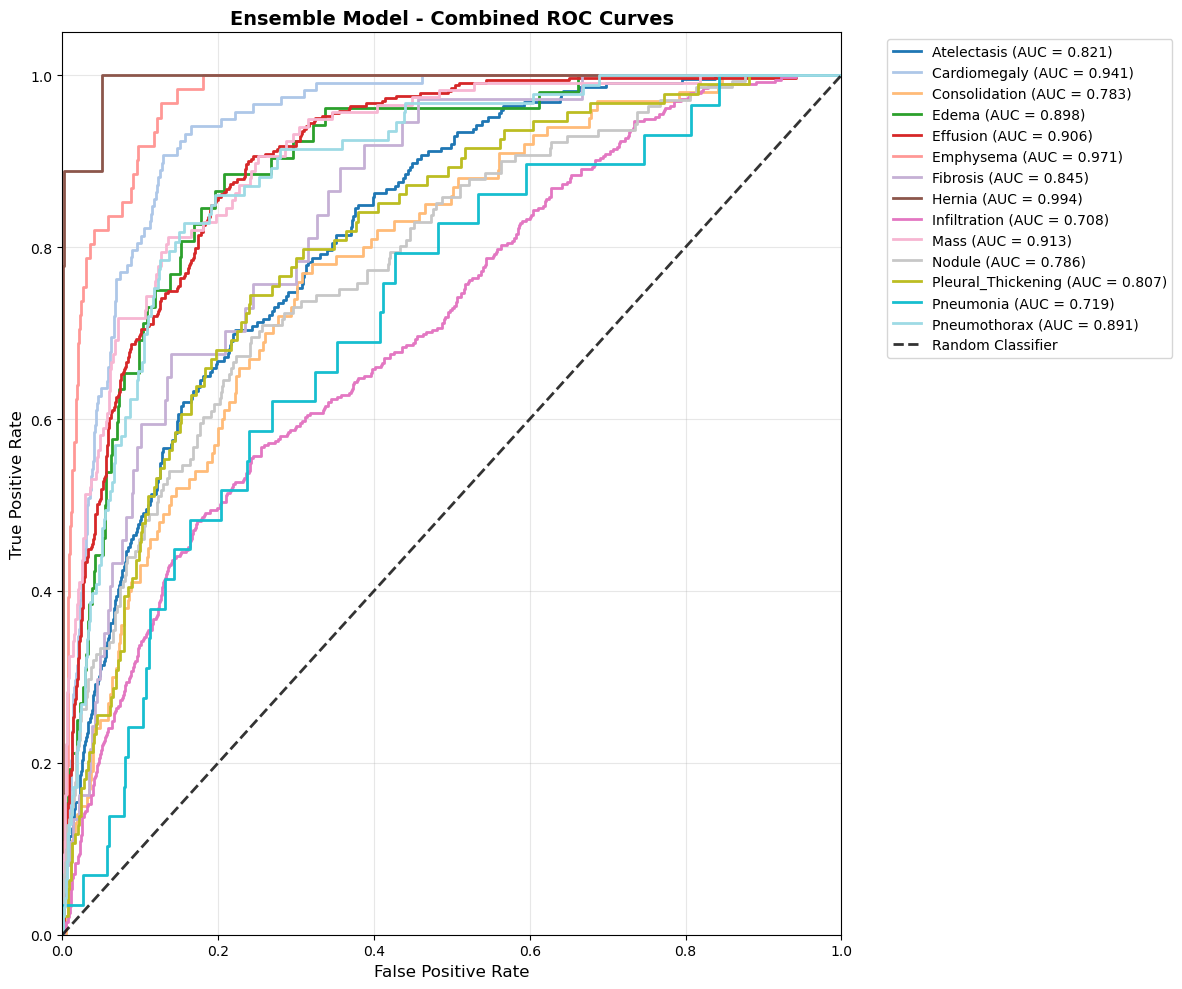}
    \caption{Ensemble Model - Combined ROC Curves}
    \label{fig:roc_curves}
\end{figure}

The synthesis of diverse models into the final Deep Ensemble successfully overcame these bottlenecks, achieving superior metrics compared to both the reproduced DannyNet baseline and the original paper’s reported results. The DE achieved an average AUROC of 0.8559, representing a consistent improvement of 0.0045 (0.53\%) over the best single model (0.8514). This gain confirms the architectural validity of the DE approach. The combined ROC curves, presented in Figure \ref{fig:roc_curves}, provide a visual validation of the robust classification performance, illustrating the strong true positive rates relative to false positive rates across all 14 classes.

\subsection{Analysis of Predictive Reliability and Uncertainty Decomposition}
The primary advantage of the Deep Ensemble is its superior quantification of uncertainty and calibration. The mean Expected Calibration Error (ECE) for the DE was 0.0728, the Mean Negative Log-Likelihood (NLL) was 0.1916, and the Mean Brier Score was 0.0478 (Table \ref{tab:Summary_of_Mean_Uncertainty_and_Calibration_Metrics_for_Deep_Ensemble}).

The dramatic reduction in all three calibration metrics confirms the success of the Deep Ensemble approach in addressing the critical requirement for trustworthy probability estimates in medical AI:

\begin{itemize}
    \item The ECE was reduced from 0.7587 to 0.0728, indicating the Deep Ensemble is highly calibrated and not prone to the catastrophic overconfidence seen in the MCD trial.
    \item The NLL decreased from 0.2526 to 0.1916, demonstrating that the DE produces probability estimates that are mathematically closer to the true labels, penalizing incorrect and overconfident predictions more effectively.
    \item The Brier Score improved from 0.0631 to 0.0478, quantifying the overall reduction in mean squared difference between predicted probabilities and actual binary outcomes, further confirming superior reliability.
\end{itemize}

\begin{figure}[!h]
    \centering
    \includegraphics[width=\columnwidth,height=0.8\textheight,keepaspectratio]{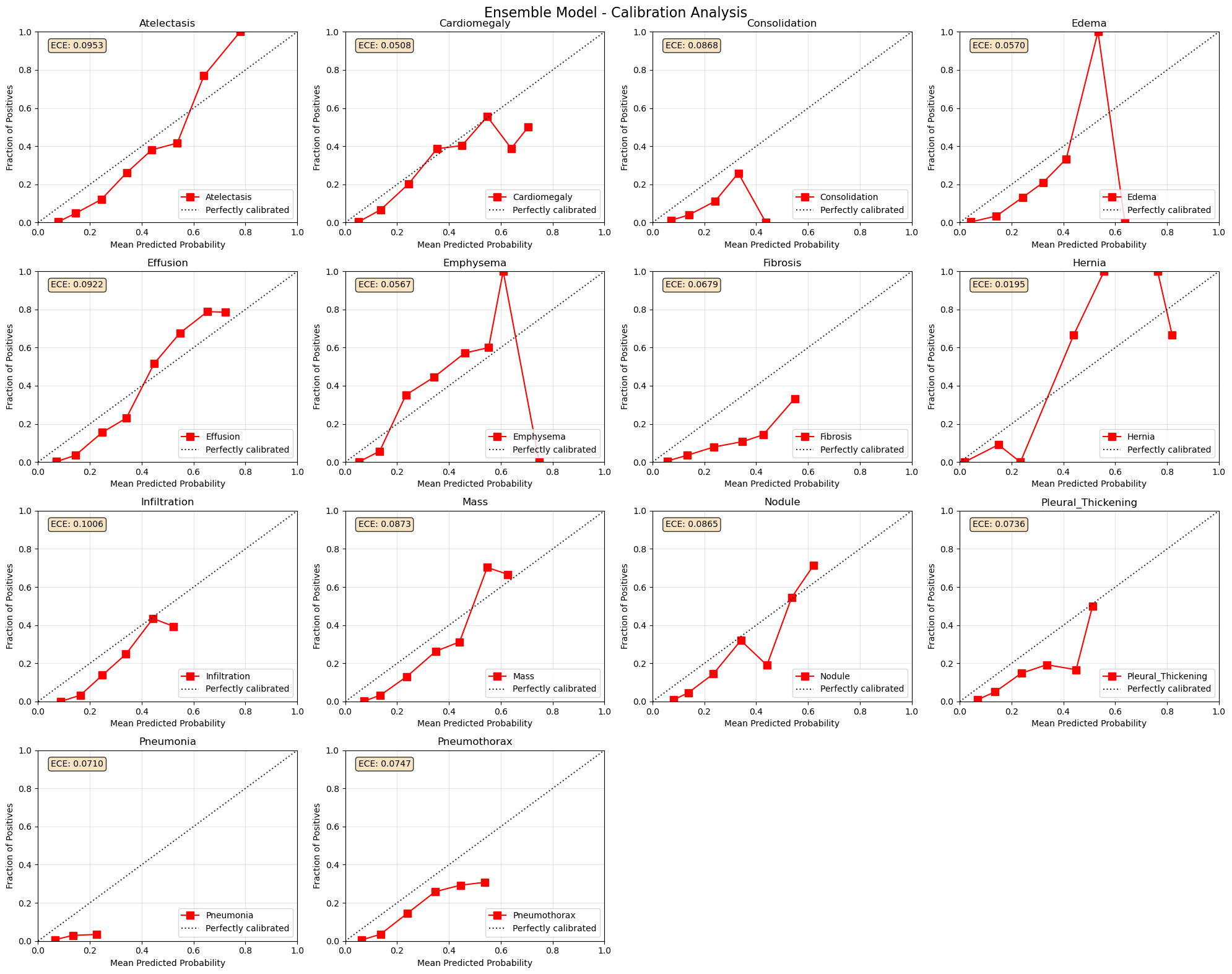}
    \caption{Ensemble Model - Calibration Analysis (ECE)}
    \label{fig:calibration_analysis_ece}
\end{figure}

This superior reliability is visualized in the ensemble calibration plots (Figure \ref{fig:calibration_analysis_ece}), which show that the predicted confidence aligns closely with the actual fraction of positives across all 14 thoracic pathologies, confirming the trustworthiness of the probability estimates.

The decomposition of uncertainty reveals a critical insight into the inherent limits of prediction in the NIH ChestX-ray14 domain: Aleatoric Uncertainty (AU = 0.3073) is overwhelmingly dominant, measuring nearly an order of magnitude larger than Epistemic Uncertainty (EU = 0.0240).   

This suggests that the Deep Ensemble has largely converged to a consistent, near optimal solution for the existing data distribution, minimizing model disagreement (Epistemic Uncertainty). Consequently, the remaining Total Uncertainty is driven primarily by irreducible data noise, such as image artifacts or inherent ambiguity in the NLP extracted labels. This finding implies that future efforts to significantly improve performance must shift focus away from purely architectural modifications and instead prioritize acquiring cleaner, clinician validated data to reduce Aleatoric Uncertainty.

\subsection{Detailed Per-Class Results and Uncertainty Correlation}
The overall metrics mask significant heterogeneity in performance and uncertainty across the 14 pathologies, reflecting the clinical variability of the diseases and the characteristics of the dataset.

As shown in Table \ref{tab:detailed_per_class_results}, a clear inverse correlation exists between classification performance (AUROC) and inherent data uncertainty (AU).

\begin{itemize}
\item \textbf{Low Uncertainty/High Performance:} Diseases with highly localized and distinct features, such as Hernia (AUROC 0.9937, AU 0.0805) and Emphysema (AUROC 0.9705, AU 0.2165), exhibit the lowest Aleatoric Uncertainty and highest predictive accuracy. This indicates that their radiological manifestations are less ambiguous, even if they are rare.

\item \textbf{High Uncertainty/Low Performance:} Pathologies characterized by diffuse, subtle, or inherently ambiguous radiological patterns, such as Infiltration (AUROC 0.7078) and Atelectasis (AUROC 0.8215), show the highest Aleatoric Uncertainty (0.5467 and 0.4267, respectively). These diseases present genuine inherent data noise that the Deep Ensemble cannot reduce, confirming that predictive difficulty is fundamentally tied to data quality and ambiguity rather than model inadequacy.
\end{itemize}

\subsection{Interpretable Uncertainty via Ensemble GradCAM}
To enhance clinical trust and transparency, Ensemble Grad-CAM visualization was implemented. This technique generates attribution heatmaps showing the regions of the chest X-ray image that contributed most strongly to the ensemble's final prediction for a specific pathology.

\begin{figure}[h]
    \centering
    \includegraphics[width=\columnwidth,height=0.25\textheight,keepaspectratio]{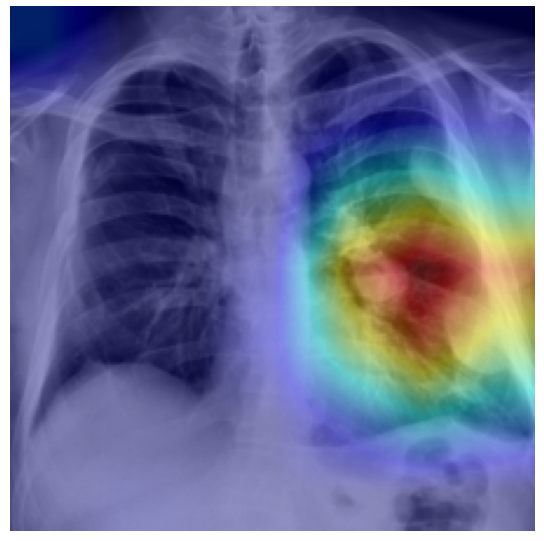}
    \caption{Example Ensemble Grad-CAM Visualization for Mass disease prediction}
    \label{fig:gradcam_visualization}
\end{figure}

Crucially, individual Grad-CAM heatmaps are computed for each of the 9 ensemble members and then averaged to produce a stable, consensus driven ensemble visualization. This collective attribution provides greater reliability than a single model Grad-CAM output. The visualization focuses on classes where the final ensemble prediction probability exceeds 0.5, highlighting regions relevant to high confidence diagnoses. This capability allows clinicians to interpret which anatomical structures the diverse set of models focused on, and in cases of high Epistemic Uncertainty, enables visual comparison of attention patterns to understand where and why models disagreed on feature relevance.

\subsection{Experimental Scope and Generalizability Constraints}
A key limitation of the current experimental phase is the exclusive reliance on the NIH ChestX-ray14 dataset for all training, validation, and final testing. While this dataset is large and representative, the performance metrics reported here are intrinsically tied to its specific data distribution and labeling conventions. Rigorous external validation on large-scale datasets such as CheXpert and MIMIC-CXR, which are crucial for confirming model generalizability and robustness to domain shift, was precluded by {restrictive data access and licensing requirements}. Therefore, the generalizability metrics
(Epistemic Uncertainty) must be interpreted within the context
of the source data distribution, necessitating external validation
as a critical next step in future work (Section \ref{sec:future_work}).

\section{Conclusion}

This research established a robust, uncertainty aware framework for multi-label chest X-ray diagnosis, effectively addressing key limitations arising from CheXNet’s reproducibility challenges and the failure of Monte Carlo Dropout to preserve classification integrity and calibration. The strategic pivot to a high diversity, nine-member Deep Ensemble was critical, yielding state-of-the-art performance (AUROC = 0.8559) and exceptional predictive reliability (ECE = 0.0728).

The validated Uncertainty Quantification (UQ) framework revealed a pivotal diagnostic insight that the primary limitation to further accuracy gains lies not in the model architecture itself but in the dominance of Aleatoric Uncertainty (AU = 0.3073), driven by data noise and labeling ambiguity. By quantifying this irreducible uncertainty, the Deep Ensemble transforms from a purely predictive model into a clinically interpretable tool that can transparently communicate its confidence and flag inherently ambiguous cases for expert review.

\section{Future Work}
\label{sec:future_work}

Given the observed dominance of Aleatoric Uncertainty and reliance on a single institutional dataset, future work should concentrate on three major directions.

\textbf{External Validation:}
A rigorous evaluation of the Deep Ensemble’s predictive and uncertainty performance on external, out-of-distribution (OOD) datasets such as CheXpert and MIMIC-CXR is essential. These datasets, derived from distinct institutions and acquisition protocols, will enable assessment of domain generalizability and validate the reliability of Epistemic Uncertainty as an indicator of data shift.

\textbf{Deeper Explainability Correlation:}
Future studies should quantitatively correlate Epistemic Uncertainty scores with visual disagreement across ensemble Grad-CAM heatmaps. Establishing a measurable relationship between elevated EU and higher inter-model visual variance would strengthen the link between uncertainty and semantic ambiguity, thereby reinforcing the interpretability of the proposed framework.

\textbf{Adaptive Ensembling:}
If an independent meta-validation set becomes available, adaptive ensemble techniques such as weighted averaging or stacking based on validation loss should be explored to further optimize predictive calibration beyond the current simple averaging strategy.

\textbf{Performance Optimization for Minority Classes:} Pursue targeted optimization strategies to improve the average F1 score across all 14 diseases. This optimization should focus particularly on classes exhibiting low F1 performance (e.g., Pneumonia, Fibrosis), potentially through further adjusting loss function weights or implementing advanced minority class resampling techniques to overcome dataset imbalance.

\section*{Data and Code Availability}
The NIH ChestX-ray14 dataset used for training, validation, and testing in this study is publicly available on Kaggle at: \url{https://www.kaggle.com/datasets/nih-chest-xrays/data}.

The complete reproducible source code for our Deep Ensemble implementation, including data preprocessing, experimental notebooks, model definitions, training procedures, uncertainty quantification, and visualization logic, is publicly accessible in our GitHub repository: \url{https://github.com/aaivu/In21-S7-CS4681-AML-Research-Projects/tree/843c66fc1b049eb1c5b79e31a333701d8ad0c141/projects/210329E-Healthcare-AI_Medical-Imaging}.


\end{document}